\documentclass[conference]{IEEEtran}
\IEEEoverridecommandlockouts

\usepackage{amsmath,amssymb,amsfonts}
\usepackage{algorithmic}
\usepackage{graphicx}
\usepackage{textcomp}
\usepackage{xcolor}
\usepackage{orcidlink}
\usepackage{adjustbox}
\usepackage{multirow}
\usepackage{makecell}
\usepackage{booktabs}

\DeclareMathOperator*{\argmin}{arg\,min}

\def\BibTeX{{\rm B\kern-.05em{\sc i\kern-.025em b}\kern-.08em
    T\kern-.1667em\lower.7ex\hbox{E}\kern-.125emX}}

\begin{document}

\title{A Unified Neural-Aided Alignment and\\ Calibration Method for AUVs}

\author{\IEEEauthorblockN{Guy Damari\,\orcidlink{0009-0001-6394-6026}\IEEEauthorrefmark{1}, \IEEEmembership{Graduate Student Member,~IEEE}, Zeev Yampolsky\,\orcidlink{0009-0003-9122-7576}, \IEEEmembership{Graduate Student Member,~IEEE},\\ and Itzik Klein\,\orcidlink{0000-0001-7846-0654}, \IEEEmembership{Senior Member,~IEEE}}
\IEEEauthorblockA{The Hatter Department of Marine Technologies \\
Charney School of Marine Sciences, University of Haifa\\
Haifa, Israel}
\thanks{\IEEEauthorrefmark{1}Corresponding author: G. Damari (email: gdamari@campus.haifa.ac.il).}}

\maketitle
\noindent
\begin{abstract}
Autonomous underwater vehicles (AUVs) rely on the fusion of inertial navigation systems (INS) and Doppler velocity logs (DVL) for accurate navigation. Before deployment, this fusion requires a DVL initialization pipeline consisting of two stages: alignment, which estimates the rotation between the INS and DVL frames, and calibration, which estimates the DVL error terms. Conventionally, both stages are solved with model-based algorithms that demand complex vehicle maneuvers, surface-level satellite reference measurements, and simplified error models, making initialization time-consuming, trajectory-dependent, and sensitive to sensor quality. In this work, we propose a fully neural-aided DVL initialization pipeline that replaces both stages with two complementary neural networks: ResAlignNet for alignment and DCNet for calibration. The unified pipeline operates in situ on a single nearly constant-velocity trajectory and uses the same inputs as the model-based baseline. Using real-world data recorded across five distinct sensor error-term combinations, the proposed pipeline reduces the velocity root mean squared error by an average of $68.7\%$ over the model-based baseline, using only $25$s of data for initialization.
\end{abstract}
\noindent

\section{Introduction}\label{sec:intro}
\noindent
Autonomous underwater vehicles (AUVs) are central to marine research, industry, and exploration~\cite{kinsey2006survey}, supporting applications such as marine geoscience and seafloor mapping~\cite{wynn2014autonomous} and the inspection of underwater structures~\cite{jacobi2015autonomous}. The success of their missions depends on the accuracy and reliability of their navigation systems~\cite{stutters2008navigation,kinsey2006survey}. Since global navigation satellite systems (GNSS) are unavailable underwater, AUVs rely on onboard sensing. The inertial navigation system (INS) provides high-rate position, velocity, and orientation from accelerometer and gyroscope measurements but drifts over time~\cite{titterton2004strapdown,groves2015principles}, while the Doppler velocity log (DVL) provides drift-free velocity relative to the seafloor from four acoustic beams~\cite{brokloff1994matrix,farrell2008aided,cohen2022beamsnet}. Their fusion has become one of the most effective solutions for precise long-range underwater navigation.
\noindent
The effectiveness of INS/DVL fusion depends on a proper DVL initialization performed before the mission. This initialization comprises two consecutive stages. The first is \textit{alignment}, which estimates the rotation between the body frame, defined by the sensitive axes of the inertial sensors, and the DVL frame; even small misalignments accumulate into large navigation errors~\cite{troni2012field}. The second is \textit{calibration}, which estimates the DVL deterministic error terms, namely the scale factor and bias, that corrupt the velocity measurement. Together, these two stages form the DVL initialization pipeline.
\noindent
In standard practice, both stages are solved with model-based algorithms that impose a significant operational burden. Alignment is commonly formulated as Wahba's problem~\cite{wahba1965least} and solved with a singular value decomposition (SVD)~\cite{umeyama2002least}, which requires specific excitation maneuvers and long convergence times~\cite{troni2012field,troni2010new}, or external positioning infrastructure such as acoustic beacons~\cite{kinsey2007in,zhaopeng2011online}. Calibration is typically performed at the surface using GNSS reference velocities and assumes a simplified scalar scale-factor model applied equally to all velocity axes~\cite{xu2020novel,liu2022gnss}; more elaborate model-based calibration employs invariant Kalman filtering~\cite{xu2022novel} or moving-base alignment~\cite{ning2023research}. As a result, the conventional process is time-consuming, trajectory-dependent, and sensitive to sensor quality, which limits initialization accuracy and prolongs mission preparation.
\noindent
Recent progress in deep learning has enabled neural-aided alternatives for underwater navigation, including DVL velocity enhancement~\cite{cohen2022beamsnet,yang2026exploring}, missing-beam reconstruction~\cite{yona2024missbeamnet,jia2026attention}, and INS/DVL fusion with uncertainty estimation~\cite{cohen2025adaptive,levy2026adaptive}. Within this trend, our prior work addressed the two initialization stages separately. ResAlignNet~\cite{damari2026resalignnet} solves the alignment stage with a one-dimensional residual network that directly regresses the alignment angles from synchronized INS and DVL velocities, operating in situ without external infrastructure or prescribed maneuvers. DCNet~\cite{yampolsky2025dcnet} solves the calibration stage with a multi-head network that estimates the DVL error terms on a nearly constant-velocity trajectory, reducing calibration time relative to model-based filters. However, these two networks were developed and evaluated in isolation; their combination into a single end-to-end pipeline, and a joint assessment of that pipeline against the full model-based initialization on real data, have not been reported.
\noindent
To bridge this gap, we propose a unified, fully neural-aided DVL initialization pipeline that chains ResAlignNet and DCNet. The main contributions are:
\begin{enumerate}
\item \textbf{End-to-end model-free initialization:} both the alignment and calibration stages are learned directly from data, removing the reliance on analytical vehicle models, prescribed maneuvers, and external infrastructure.
\item \textbf{Simplified trajectory requirement:} the entire initialization is performed on a single nearly constant-velocity trajectory, substantially reducing operational complexity.
\end{enumerate}
\noindent
We evaluate the proposed pipeline end-to-end against the full model-based baseline on real AUV data over five distinct error-term combinations, isolating the marginal contribution of the calibration stage through an ablation. Using only $25$s of data for initialization, the proposed pipeline reduces the average velocity root mean squared error (VRMSE) by $68.7\%$ over the baseline. This substantial error reduction from only a short, maneuver-free segment enables faster, simpler, and more reliable AUV deployment, directly lowering the operational cost of accurate underwater navigation.

\noindent
The remainder of this paper is organized as follows. Section~\ref{sec:problem} formulates the DVL velocity, INS, alignment, and calibration models. Section~\ref{sec:approach} presents the proposed unified pipeline. Section~\ref{sec:results} reports the experimental validation, and Section~\ref{sec:conclusions} concludes the paper.

\section{Problem Formulation}\label{sec:problem}

\subsection{INS/DVL Alignment}\label{subsec:align}
\noindent
The alignment stage estimates the rotation $\mathbf{T}^{d}_{b}\in SO(3)$ between the body frame and the DVL frame. The standard in-situ approach integrates the inertial acceleration to form a body-frame velocity and matches it to the DVL velocity~\cite{troni2012field}, casting the problem as Wahba's problem~\cite{wahba1965least}. The optimal rotation is obtained by minimizing the velocity residual over $N$ time instances,
\begin{equation}\label{eq:wahba}
\hat{\mathbf{T}}^{d}_{b}=\argmin_{\mathbf{T}\in SO(3)}\frac{1}{N}\sum_{i=1}^{N}\left\lVert \boldsymbol{v}^{b}[t_i]-\mathbf{T}_{d}^{b}\tilde{\boldsymbol{v}}^{d}[t_i]\right\rVert^{2}
\end{equation}
where $\boldsymbol{v}^{b}$ is the body-frame velocity obtained by integrating the estimated inertial acceleration, $\tilde{\boldsymbol{v}}^{d}$ is the DVL velocity measured in the DVL frame, and $\mathbf{T}_{d}^{b}$ is the rotation that maps the DVL velocity into the body frame. Eq.~\eqref{eq:wahba} is solved with an SVD algorithm~\cite{umeyama2002least}. This baseline is accurate only when the trajectory excites all axes sufficiently and enough data is collected, and it degrades markedly with lower-grade sensors.

\subsection{DVL Calibration Error Model}\label{subsec:calib}
\noindent
After alignment, calibration estimates the deterministic DVL error terms relating the measured DVL velocity to a reference velocity. We adopt the comprehensive six-term model~\cite{yampolsky2025dcnet}
\begin{equation}\label{eq:error_model}
\tilde{\boldsymbol{v}}^{d} = (1+\boldsymbol{k}_{\mathrm{DVL}})\,\hat{\mathbf{T}}^{d}_{b}\,
\boldsymbol{v}^{b}_{\mathrm{GNSS}} + \boldsymbol{b}_{\mathrm{DVL}} + \boldsymbol{\delta v}^{d}
\end{equation}
where $\boldsymbol{v}^{b}_{\mathrm{GNSS}}$ is the reference velocity, $\boldsymbol{\delta v}^{d}$ is zero-mean Gaussian white noise, and $\boldsymbol{k}_{\mathrm{DVL}}$ and $\boldsymbol{b}_{\mathrm{DVL}}$ are the DVL scale-factor and bias vectors, respectively. The unknowns are the alignment $\hat{\mathbf{T}}^{d}_{b}$, the scale-factor vector $\boldsymbol{k}_{\mathrm{DVL}}\in\mathbb{R}^{3}$, and the bias vector $\boldsymbol{b}_{\mathrm{DVL}}\in\mathbb{R}^{3}$. Unlike the scalar scale-factor baseline~\cite{xu2020novel,liu2022gnss}, which applies a single scale to all axes, \eqref{eq:error_model} allows an independent scale and bias per axis, offering the flexibility required for precise per-axis correction.

\section{Proposed Unified Approach}\label{sec:approach}
\begin{figure*}[t]
    \centering
    \begin{adjustbox}{width=0.85\textwidth}
    \includegraphics[width=0.99\linewidth]{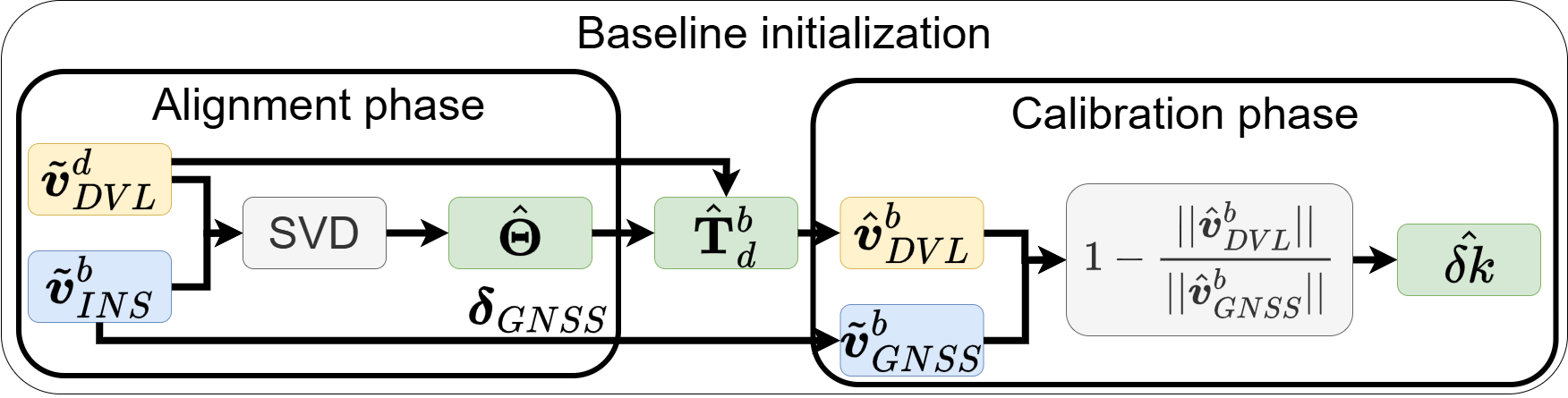}
    \end{adjustbox}
    \caption{Conventional model-based DVL initialization pipeline, comprising the SVD-based alignment phase and the scalar scale-factor calibration phase.}
    \label{fig:baseline_pipeline}
\end{figure*}

\begin{figure*}[t]
    \centering
    \begin{adjustbox}{width=0.85\textwidth}
    \includegraphics[width=0.99\linewidth]{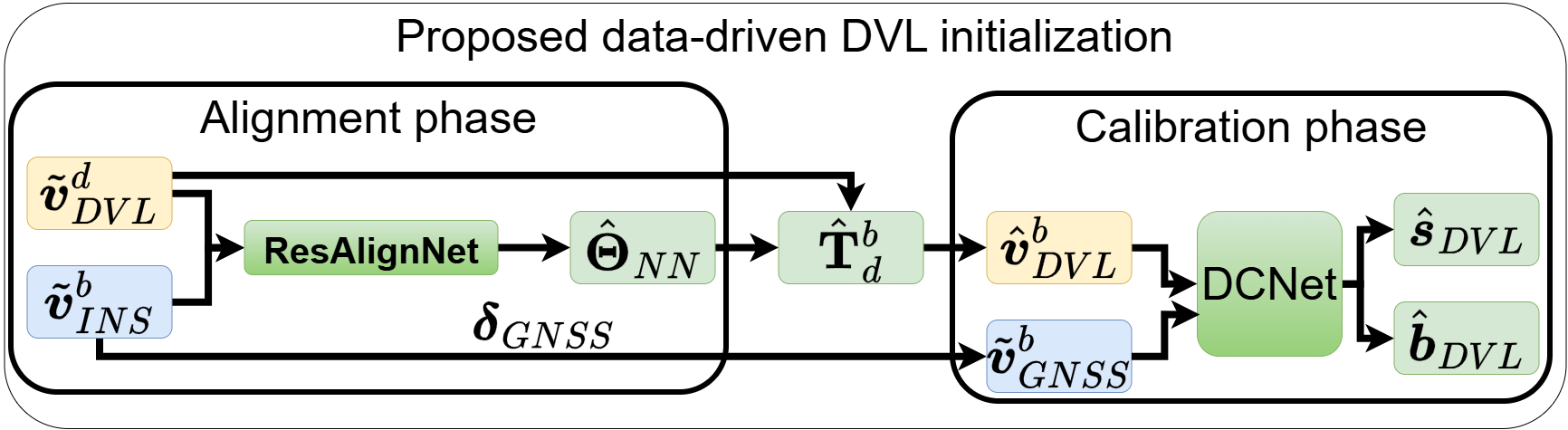}
    \end{adjustbox}
    \caption{Proposed neural-aided DVL initialization pipeline: ResAlignNet performs the alignment phase, and DCNet performs the calibration phase.}
    \label{fig:prop_pipeline}
\end{figure*}

\noindent
The conventional model-based initialization pipeline, comprising SVD-based alignment followed by scalar scale-factor calibration, is illustrated in Fig.~\ref{fig:baseline_pipeline}. To enhance the navigation permanence, we offer to replace both stages with two complementary neural networks while keeping the same inputs. Specifically, we employ ResAlignNet to estimated the body to DVL frame transformation matrix, and DCNet to estimate the DVL scale and bias error terms. The proposed pipeline is shown in Fig.~\ref{fig:prop_pipeline}.

\subsection{Alignment Stage: ResAlignNet}\label{subsec:resalignnet}
\noindent
ResAlignNet~\cite{damari2026resalignnet} is a one-dimensional ResNet-18 architecture~\cite{he2016deep} that transforms the alignment problem into a supervised regression task. The input is the concatenation of the synchronized INS and DVL velocity windows, forming a tensor of size $[\textit{window\_size}\times 6]$, where the six channels are the three velocity components of each sensor. An initial wide convolution is followed by four residual layers with increasing channel dimensions ($64,128,256,512$); the skip connections preserve gradient flow and enable robust feature extraction under sensor noise. A global average pooling layer condenses the temporal dimension, and a fully connected layer outputs the three Euler angles $(\phi,\theta,\psi)$ that define $\hat{\mathbf{T}}^{d}_{b}$. The network is trained by minimizing the mean squared error (MSE) between the predicted and ground-truth (GT) angles,
\begin{equation}\label{eq:mse_align}
\mathrm{MSE}=\frac{1}{N}\sum_{i=1}^{N}\sum_{j\in\{\phi,\theta,\psi\}}\left(\alpha_{i,j}-\hat{\alpha}_{i,j}\right)^{2}
\end{equation}
using the Adam optimizer. Because ResAlignNet operates directly on velocity windows, it requires neither prescribed maneuvers nor external positioning, and it converges within seconds of data.

\subsection{Calibration Stage: DCNet}\label{subsec:dcnet}
\noindent
DCNet~\cite{yampolsky2025dcnet} is a multi-head network that estimates the DVL error terms from the aligned DVL and reference velocities expressed in the body frame. The input, of size $[6\times\textit{window\_size}]$, stacks the DVL and GNSS velocity axes. A 1D convolutional head processes the element-wise difference of the two velocity vectors, which under the error model isolates the scale contribution, while a 2D convolutional head with a dilated kernel jointly processes the corresponding axes of both vectors. The two heads are concatenated and passed through a fully connected batch that outputs the error terms. We employ the six-term vector error model of~\eqref{eq:error_model}, which estimates an independent scale factor and bias per axis and yields the most precise correction. The network is trained with an MSE loss computed on the calibrated velocity rather than on the error terms themselves,
\begin{equation}\label{eq:mse_calib}
\mathrm{MSE}=\frac{1}{N}\sum_{i=1}^{N}\sum_{j\in\{x,y,z\}}\left(\boldsymbol{y}_{i,j}-\hat{\boldsymbol{y}}_{i,j}\right)^{2}
\end{equation}
where $\boldsymbol{y}$ and $\hat{\boldsymbol{y}}$ are the reference and calibrated velocities.

\subsection{End-to-End Pipeline}\label{subsec:pipeline}
\noindent
The two stages are chained into a single initialization procedure. Given a calibration trajectory, ResAlignNet first estimates the alignment $\hat{\mathbf{T}}^{d}_{b}$, which is used to transform the DVL measurements into the body frame. The aligned DVL and reference velocities are then passed to DCNet, which estimates the per-axis scale and bias. The resulting alignment and error terms constitute the complete DVL initialization and are subsequently applied to correct any evaluation trajectory. Crucially, both stages consume only velocity windows and therefore operate on a single nearly constant-velocity trajectory, eliminating the maneuver and surface-reference requirements of the model-based pipeline.

\section{Experimental Results}\label{sec:results}

\subsection{Dataset}\label{subsec:dataset}
\noindent
We validate our proposed pipeline using the A-KIT dataset~\cite{cohen2025adaptive}, employing six trajectories totaling $33$ minutes: one for network training, one for calibration, and four for evaluation. All trajectories were collected during real-world sea trials and include synchronized DVL and inertial measurements, along with a full navigation solution. The calibration and evaluation trajectories are further detailed in Section~\ref{subsec:procedure}. Since Snapir carries a single high-end DVL, a controlled noising pipeline is applied to the recorded measurements to emulate different DVL grades through different error-term combinations. ResAlignNet and DCNet are trained on nearly constant-velocity trajectories noised with multiple error-term combinations, while ensuring that the evaluation combinations are excluded from training.

\subsection{Validation Procedure}\label{subsec:procedure}
\begin{figure}[t]
    \centering
    \includegraphics[width=0.99\linewidth]{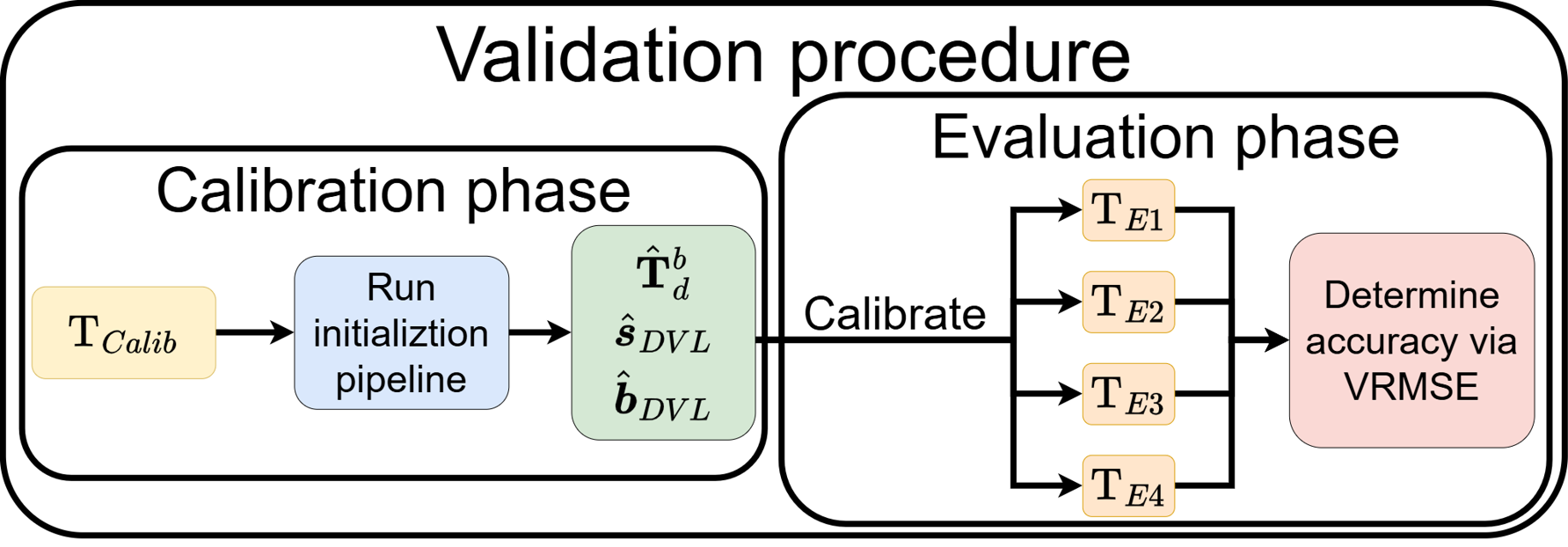}
    \caption{Validation procedure. In the calibration phase, the initialization pipeline is run on $T_{\mathrm{Calib}}$ to estimate the alignment and DVL error terms; in the evaluation phase, four independent trajectories are calibrated and assessed via VRMSE.}
    \label{fig:validation}
\end{figure}

\noindent
The validation procedure comprises two consecutive phases, illustrated in Fig.~\ref{fig:validation}. In the \textit{calibration phase}, a single calibration trajectory $T_{\mathrm{Calib}}$ is used to run the full initialization pipeline and estimate the alignment $\hat{\mathbf{T}}^{d}_{b}$ and the DVL error terms $\hat{\boldsymbol{s}}_{\mathrm{DVL}},\hat{\boldsymbol{b}}_{\mathrm{DVL}}$. In the \textit{evaluation phase}, these estimates are applied to four independent evaluation trajectories $T_{E1}$--$T_{E4}$, which are transformed to the body frame and calibrated. For a trajectory of $N$ velocity samples, the VRMSE against the GT is
\begin{equation}\label{eq:vrmse}
\mathrm{VRMSE}=\sqrt{\frac{1}{N}\sum_{i=1}^{N}\left\lVert \boldsymbol{v}_{i}-\hat{\boldsymbol{v}}_{i}\right\rVert^{2}}
\end{equation}
where $\boldsymbol{v}_{i}$ and $\hat{\boldsymbol{v}}_{i}$ are the GT and estimated velocity vectors at sample $i$, respectively. The VRMSE is computed for each evaluation trajectory and averaged. The entire procedure is repeated five times, each with a different error-term combination (ETC1--ETC5). All results reported below use a $25$s initialization segment. The five combinations, spanning a range of alignment errors, per-axis scale factors, biases, and noise levels, are summarized in Table~\ref{tbl:etc}. For each ETC, the procedure is executed for both the model-based baseline, consisting of SVD alignment~\cite{umeyama2002least} and scalar scale-factor estimation~\cite{liu2022gnss}, and the proposed neural-aided pipeline.

\begin{table}[t]
\centering
\caption{Evaluation error-term combinations (ETC1--ETC5). Alignment errors are per-axis Euler angles; scale factor and bias are per-axis DVL error terms.}
\label{tbl:etc}
\scriptsize
\setlength{\tabcolsep}{2pt}
\begin{tabular}{@{}lccccc@{}}
\toprule
 & \makecell{Euler\\{[deg]}} & \makecell{Scale\\factor [\%]} & \makecell{DVL bias\\{[m/s]}} & \makecell{DVL STD\\{[m/s]}} & \makecell{GNSS STD\\{[m/s]}} \\
\midrule
ETC1 & [5, 4.5, 3.6] & [0.5, 0.1, 0.2] & [.0005,.0002,.0001,.0005] & 0.0050 & 0.005 \\
ETC2 & [4.2, 5, 3]   & [0.2, 0.1, 0.3] & [.0003,.0001,.0002,.0005] & 0.0030 & 0.005 \\
ETC3 & [3, 4.2, 0.2] & [0.2, 0.3, 0.4] & [.0004,.0002,.0001,.0003] & 0.0001 & 0.005 \\
ETC4 & [3.5, 4.5, 5] & [0.4, 0.1, 0.2] & [.0006,.0004,.0005,.0001] & 0.0020 & 0.005 \\
ETC5 & [0.5, 5, 4.6] & [0.2, 0.5, 0.4] & [.0001,.0002,.0003,.0005] & 0.0040 & 0.005 \\
\bottomrule
\end{tabular}
\end{table}

\subsection{Accuracy Comparison}\label{subsec:accuracy}
\begin{figure}[t]
    \centering
    \includegraphics[width=0.9\linewidth]{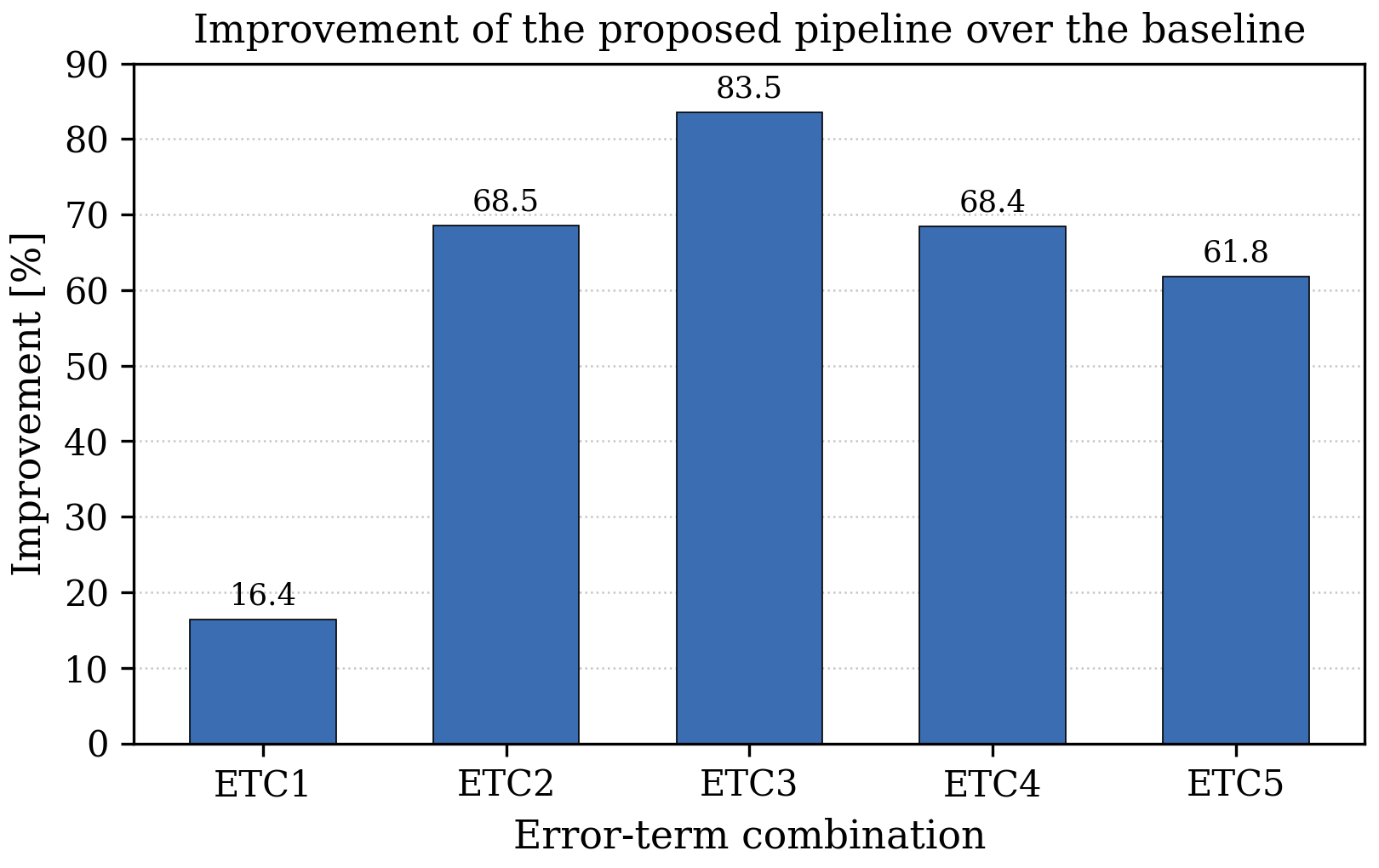}
    \caption{Improvement in average VRMSE of the proposed pipeline over the baseline for each error-term combination, using a $25$s initialization segment.}
    \label{fig:improvement}
\end{figure}

\noindent
Table~\ref{tbl:results} reports the average VRMSE over the four evaluation trajectories for each ETC, for both the baseline and the proposed pipeline, using a $25$s initialization segment. The proposed pipeline consistently outperforms the baseline across all five combinations, reducing the average VRMSE from $0.5768$m/s to $0.1803$m/s. Fig.~\ref{fig:improvement} shows the per-combination improvement, ranging from $16.4\%$ to $83.5\%$, with an average improvement of $68.7\%$ computed from the averaged VRMSE. The baseline is particularly sensitive to the more aggressive combinations (e.g., ETC3), where its VRMSE approaches $1.0$m/s, whereas the proposed pipeline remains below $0.21$m/s in every case.

\begin{table}[t]
\centering
\caption{Average VRMSE [m/s] over the four evaluation trajectories using a $25$s initialization segment, for each error-term combination. In bold is the improvement of the proposed pipeline over the baseline.}
\label{tbl:results}
\scriptsize
\setlength{\tabcolsep}{3pt}
\begin{tabular}{@{}lcc@{}}
\toprule
 & Baseline [m/s] & Ours [m/s] ([\%]) \\
\midrule
ETC1    & 0.2018 & 0.1687 (\textbf{16.4}) \\
ETC2    & 0.5933 & 0.1866 (\textbf{68.5}) \\
ETC3    & 0.9843 & 0.1623 (\textbf{83.5}) \\
ETC4    & 0.5791 & 0.1831 (\textbf{68.4}) \\
ETC5    & 0.5253 & 0.2008 (\textbf{61.8}) \\
\midrule
Average & 0.5768 & 0.1803 (\textbf{68.7}) \\
\bottomrule
\end{tabular}
\end{table}

\subsection{Ablation: Contribution of the Calibration Stage}\label{subsec:ablation}
\noindent
To isolate the marginal contribution of the calibration stage, we compare the VRMSE obtained when the pipeline is stopped after alignment (ResAlignNet only) with that obtained when DCNet is subsequently applied (ResAlignNet~$+$~DCNet). Table~\ref{tbl:ablation} reports both, per error-term combination. The alignment stage alone already reduces the VRMSE far below the baseline, confirming that misalignment is the dominant error source in DVL initialization. Adding DCNet yields a further, smaller reduction, lowering the average VRMSE from $0.1843$m/s to $0.1803$m/s (a $2.2\%$ reduction) and improving four of the five combinations by up to $7.4\%$, while marginally increasing the error for ETC3. This indicates that once the alignment is accurate, the residual DVL scale and bias errors are small, so the calibration stage provides a modest refinement.

\begin{table}[t]
\centering
\caption{Ablation study: average VRMSE [m/s] with ResAlignNet only versus the full ResAlignNet~$+$~DCNet pipeline, per error-term combination.}
\label{tbl:ablation}
\scriptsize
\setlength{\tabcolsep}{4pt}
\begin{tabular}{@{}lcc@{}}
\toprule
 & \makecell{ResAlignNet\\only [m/s]} & \makecell{Full pipeline\\(ResAlignNet~$+$~DCNet) [m/s]} \\
\midrule
ETC1    & 0.1740 & 0.1687 \\
ETC2    & 0.1870 & 0.1866 \\
ETC3    & 0.1528 & 0.1623 \\
ETC4    & 0.1976 & 0.1831 \\
ETC5    & 0.2103 & 0.2008 \\
\midrule
Average & 0.1843 & 0.1803 \\
\bottomrule
\end{tabular}
\end{table}

\section{Conclusions}\label{sec:conclusions}
\noindent
This work presented a unified, fully neural-aided DVL initialization pipeline that chains ResAlignNet for alignment and DCNet for calibration, replacing the conventional SVD-and-scalar-scale model-based calibration procedure. Evaluated end-to-end on real Snapir AUV data across five sensor error-term combinations, the proposed pipeline reduced the average VRMSE by $68.7\%$ over the baseline using only $25$s of data for initialization. 
\noindent
The results demonstrate that a unified neural-aided pipeline can replace both stages of the conventional DVL initialization while using the same inputs. Beyond the accuracy gain, the practical advantages are threefold: 1)~the pipeline operates in situ without external positioning, 2)~it requires only a single nearly constant-velocity trajectory rather than dedicated maneuvers, and 3)~it delivers stable accuracy almost immediately, shortening mission preparation. These properties directly address recurring operational challenges in underwater navigation, where inaccurate initialization, and misalignment in particular, translates into large positioning errors that compromise both mission success and data validity. Future work will extend the evaluation to a broader range of misalignment magnitudes and DVL grades, and integrate the pipeline into a full in-mission navigation solution.

\section*{Acknowledgement}
\noindent
G. D. is supported by the Maurice Hatter Foundation and the University of Haifa excellence scholarship for Ph.D. studies. Z. Y. is supported by the Maurice Hatter Foundation and the University of Haifa presidential scholarship for outstanding students on a direct Ph.D. track.

\end{document}